\documentclass[10pt,twocolumn,letterpaper]{article}

\usepackage{graphicx}
\usepackage{subfigure}
\usepackage{booktabs}
\usepackage{amsmath}
\usepackage{amssymb}
\usepackage{bm}
\usepackage{appendix}

\newtheorem{assumption}{Assumption}
\usepackage{algorithm}
\usepackage{algorithmic}

\usepackage[breaklinks=true,bookmarks=false]{hyperref}

\begin{document}

\title{Coconditional Autoencoding Adversarial Networks for Chinese Font Feature Learning}

\author{Zhizhan Zheng\\
Hanyi Fonts\\
{\tt\small zhizhanzh@zju.edu.cn}
\and
Feiyun Zhang\\
Qiniu Atlab\\
{\tt\small zhangfeiyun@qiniu.com}
}
\date{}

\maketitle

\begin{abstract}
In this work, we propose a novel framework named Coconditional Autoencoding Adversarial Networks (CocoAAN) for Chinese font learning, which jointly learns a generation network and two encoding networks of different feature domains using an adversarial process. The encoding networks map the glyph images into style and content features respectively via the pairwise substitution optimization strategy, and the generation network maps these two kinds  of features to glyph samples. Together with a discriminative network conditioned on the extracted features, our framework succeeds in producing realistic-looking Chinese glyph images flexibly. 
Unlike previous models relying on the complex segmentation of Chinese components or strokes, our model can 
``parse'' structures in an unsupervised way, through which the content feature representation of each character is captured.
Experiments demonstrate our framework has a powerful generalization capacity to other unseen fonts and characters.  
\end{abstract}

\section{Introduction}
The Chinese writing system and through it a sense of a common literature and history is the very fabric that held China together for millennia. 
But unlike phonetic writing system which have very limited number of letters such as English, Chinese has a huge amount of ideographic characters(more than 80000).
On the other hand, most Chinese characters have more complex shapes and structures than other symbolic characters.
That is why designing a new Chinese font is such an expensive and difficult work, it needs a group of type designers and calligraphers working together for years for a typerface covering official character set like GB-18030, which contains 27533 unique characters.
Due to this fact, we could seldomly see independent artists working on Chinese typerface design, which also leads to the status quo that there existed fewer well-designed Chinese digital fonts than alphabetic ones.
To alleviate the labor intensive part in this design process, various automatic synthesizing approaches have been proposed these years, among them, machine learing method is a promising solution.

\begin{figure*}[htb]
  \centering
  \includegraphics[width=.8\linewidth]{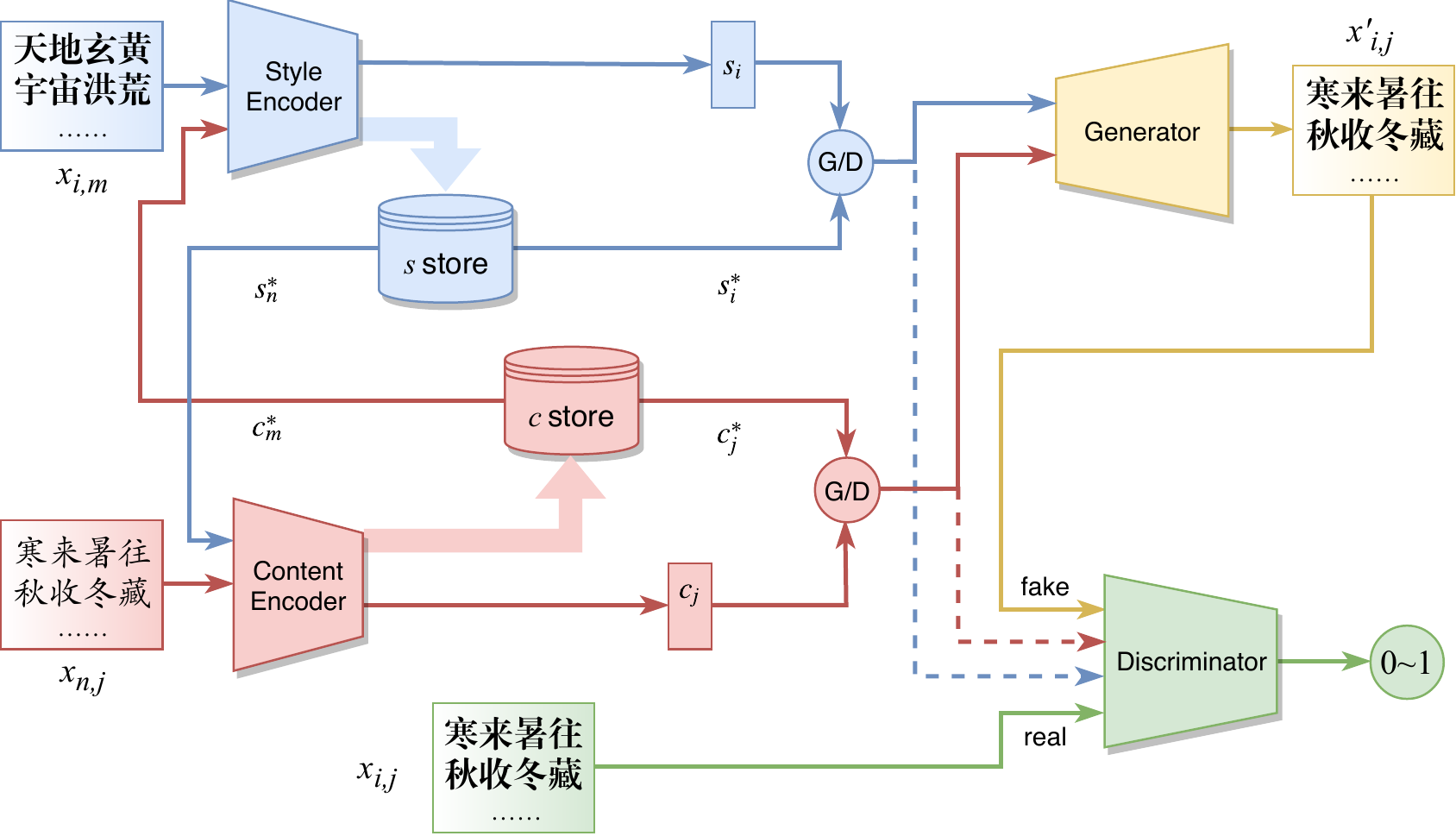}
  \caption{An overview of the process of CocoAAN. (Best viewed in colors).
  $x_{i, j}$, $x_{i, m}$ and $x_{n, j}$ denote training glyph samples of different fonts and characters, 
  $x_{i, j}'$ denote the reconstructed sample of font $i$ and character $j$,
  $\bm{s_i}$ and $\bm{c_j}$ denote the style and content representation extracted from corresponding glyph sample, while $\bm{s_i^*}$, $\bm{s_n^*}$, $\bm{c_j^*}$ and $\bm{c_m^*}$ are feature vectors feeding from the style store or content store.
  }
  \label{fig-overall}
\end{figure*}

In this work, we break the Chinese font learing task down into 2 subtasks (the encoding part and the adversarial), and fufill each with 2 convolutional networks. The encoding networks are designed to disentangle content and style features separately,  with a pairwise substitution optimization we force the networks to capture lattent embeddings of feature from different  domains. 
The overall process of our proposed framwork is shown in Fig. \ref{fig-overall}. The main contributions of this work are:
\begin{itemize}
  \item We propose a novel model which can disentangle each Chinese glyph image into content and style representations automatically, and with those two kinds of feature embedding, the model can generate specified glyph in clear appearance.
  \item We incoporate the adversarial netwroks with the auto-encoders, and propose an adversarial way to train the auto-encoders. 
  \item We demonstrate the potention of our proposed framwork to be generalized to new fonts and other characters beyond the training sets.  
\end{itemize}

\section{Related Work}\label{relatedwork}
Early researches \cite{Xu2009AutomaticGO, IAAI148259} on synthesizing Chinese characterss are mostly stroke-based methods which formulate the target character generation as a process of assembling all needed components segmented from training sets.
So one of the most important aims of their models can be sumarized as finding a suitable algorithm to decompose Chinese characters into hierarchical representation of simple radicals and strokes.
This process resembles the human being's way to learn Chinese characters' structure, as it is apparent that the Chinese characters do constitute of repeated componets.
However, the intrinsic assembling grammer is not that straightforward as it seemed, there always are some complicated characters that could  hardly be correctly decomposed.
Apart from the hevily rely on the preceding parsing, those stroke-based methods 
paid little attention on writing style transfering other than the local representations of the characters, which also prevents its application in the Chinese character generation.

Zhang et al.\cite{zhangDrawingRecognizingChinese2016} proposed a sequence-based method which uses Recurrent Neural Networks (RNN) to extract temporal information of ordinal hand-written strokes to generate skeletons of Chinese characters. 
Since the generated skeletons by this method contain only the structure appearance but no style information, it can not generalize to different style glyphs.

zi2zi\footnote{https://github.com/kaonashi-tyc/zi2zi/} and other relevant methods \cite{changGeneratingHandwrittenChinese2018,changChineseTypographyTransfer2017} implements the Chinese characters’ style transfer based on pix2pix\cite{isolaImagetoImageTranslationConditional2016}, which considers the generation as a problem of mapping from the source domain to the target pairwisely. 
Theoretically, different glyphs of a same font do share the same writing style, there should be a fixed mapping for any glyphs of two uniformly designed fonts.
However, this kind of methods only concentrate on the mapping relationship in a specified font pair, while ignoring a lot more font resouces which would also lead to a lack of flexibility for new style learning, as for every new font pair, the netwroks must be retrained from scratch.

Sun et al.\cite{sunUnsupervisedTypographyTransfer2018} invoked an intercross pairwise scheme to infer the common style feature, which takes advantage of the implicit style co-sharing nature of different fonts.
But for the character content feature they used a manual encoding method based on the radical assembling knowledge of each character, which could hardly be generalized to sophisticated structure as the stroke-based methods.
Moreover, the generated samples are often blurry, which should attribute to the disadvantage of its varitional auto-encoder (VAE) mechanism as it uses the pixel-wise loss for reconstruction objective.

\section{Methodology}
As shown in Fig. \ref{fig-overall}, our proposed CocoAAN consists of two stages, the encoding part and the adversarial part, each of which consists two subnets, i.e., the former includes the style and content subnets (the blue and red trapezoids in Fig. \ref{fig-overall}) while the latter involves the generator and discriminator subnets (the yellow and green trapezoids in Fig. \ref{fig-overall}) . In this section, we will present all details of each subnet and the process of optimization. 

\subsection{Assumptions and Encoding Networks}
In contrast to the challenges mentioned above, Chinese fonts are particularly well suited for cross domain transfer learing, since they naturally are database of aligned style pairs and content pairs. 
A font is a collection of glyphs which representing different characters, from a same font, every glyph should share a specific weight, width, slant, ornamentation, i.e., they share a set of common design features, while across fonts, glyphs of same character should share a same radical assembing relationship and inherent stroke structure.

\begin{assumption}\label{asp-cs}
  Glyphs of one same character but from different fonts share a same content feature; Glyphs of different characters but from one same font share a same style feature.
\end{assumption}

Moreover, we agreed the assumption from Sun et al.\cite{sunUnsupervisedTypographyTransfer2018} that a glyph can be characterized by, and only by two sides of feature:  
\begin{assumption}\label{asp-x-z}
  Every glyph contains features from two independent domains, the style factor and the content factor, by which uniquely determined a glyph, which can be formulated as:
\end{assumption}

\begin{equation}
  x_{i,j}\rightleftharpoons (\bm{s_i}, \bm{c_j})
\end{equation}
where $x_{i, j}$ represents the glyph of font $i$ and character $j$, $\bm{s_i}$ and $\bm{c_j}$ denote the style and content feature of the specified glyph respectively.

Upon Assumption \ref{asp-x-z}, the key to solving this problem is decoupling every glyph $x_{i,j}$ into its style feature $\bm{s}$ and content feature $\bm{c}$ representations. As soon as we obtain the precise feature embeddings, we could reconstruct any glyph from existing style code $\bm{s}$ and content code $\bm{c}$ flexibly.
In our approach, we use two encoding networks to capture feature codes respectively from glyph samples.
To decisely disentangle these two hidden features, besides the glyph image $x_{i,j}$, either feature embedding $\bm{s_i}$ or $\bm{c_j}$ should also be as the input of these two networks.
Let $\mathcal{C}$ and $\mathcal{S}$ be the corresponing encoding networks, the feature extracting process can be considered as a function of $x$ and $\bm{s}$ or $\bm{c}$, then the co-sharing nature of Assumption \ref{asp-cs}  can be illustrated as:
\begin{equation}\label{eq-netc}
  \bm{c_j} = \mathcal{C}(x_{i, j}, \bm{s_i}^*) = \mathcal{C}(x_{n, j}, \bm{s_n}^*)
\end{equation}
\begin{equation}\label{eq-nets}
  \bm{s_i} = \mathcal{S}(x_{i, j}, \bm{c_j}^*) = \mathcal{S}(x_{i, m}, \bm{c_m}^*)
\end{equation}
where $\bm{s_i}^*$, $\bm{s_n}^*$, $\bm{c_j}^*$ and $\bm{c_m}^*$ denote different style or content features which were treated as a known input of the conditional networks, specifically, they were supplied from the encoding results of the previous stage of network $\mathcal{S}$ or $\mathcal{C}$.

\subsection{Adversarial Networks}
Generative adversarial networks (GAN)\cite{goodfellowGenerativeAdversarialNetworks2014} establishes a min-max adversarial game between its generattor $\mathcal{G}$ and discriminator $\mathcal{D}$: $\mathcal{G}$ produces model distribution mapped from a latent random noise, and $\mathcal{D}$ distinguishes the model distribution from the target. By consecutively training the model with the minimax objective in turn:
\begin{equation}
  \min_\mathcal{G} \max_{\mathcal{D}} \mathbb{E}_{x\sim p_{\mathrm{data}}}[\log \mathcal{D}(x)] + \mathbb{E}_{\bm{z} \sim p(\bm{z} )}[\log(1-\mathcal{D}(\mathcal{G}(\bm{z})))]
\end{equation}
it can achieve impressive results in a wide range of generative scenarios, as with its discriminator network it provides a more abstract metric for complex data distribution than other element-wise reconstruction errors.

The conditional GAN (cGAN)\cite{mirzaConditionalGenerativeAdversarial2014} is an extension of the vanilla GAN where both $\mathcal{D}$ and $\mathcal{G}$ receives an additional condition $\bm{y}$ as input, which controls the domain of the output distribution. Our apporach is based on the framework of cGAN with two kinds of conditions, but no random noise $\bm{z}$ as input.
Conditioned on $\bm{c_j}$ and $\bm{s_i}$ captured by the above-mentioned encoding networks, the adversary between generator($\mathcal{G}$) and discriminator($\mathcal{D}$) can be expressed as alternatively maxmizing $L(\mathcal{G}^*, \mathcal{D})$ in Eq. (\ref{eq-loss-D}) and minimizing $L(\mathcal{G}, \mathcal{D}^*)$ in Eq. (\ref{eq-loss-G}):
\begin{equation}\label{eq-loss-D}
  \begin{aligned}
    L(\mathcal{G}^*, \mathcal{D}) = {}  & \mathbb{E}_{x_{i, j} \sim p_{\mathrm{data}}}[\log \mathcal{D}(x_{i, j}|\bm{s_i}, \bm{c_j})] + \\
    & \mathbb{E}_{x'_{i, j}  \sim p_{\{G^*(\bm{s_i}^*, \bm{c_j}^*)\}}}[\log(1-\mathcal{D}(x'_{i, j}|\bm{s_i}, \bm{c_j}))]
  \end{aligned}
\end{equation}
\begin{equation}\label{eq-loss-G}
  L(\mathcal{G}, \mathcal{D}^*) = \mathbb{E}_{x'_{i, j}  \sim p_{\{\mathcal{G}(\bm{s_i}, \bm{c_j})\}}}[\log(1-\mathcal{D}(x'_{i, j}|\bm{s_i}^*, \bm{c_j}^*))]
\end{equation}
where $p_{\mathrm{data}}$ and $p_{\{\mathcal{G}(\bm{s_i}, \bm{c_j})\}}$ correspondingly  denotes the distribution  of true data and data generated from the given inputs $\bm{s_i}$ and $\bm{c_j}$. ``$*$'' labeled in upper right of the network's symbol indicates the network's parameters was fixed in current iteration.

By Equation (\ref{eq-loss-D}), the parameters  of $\mathcal{S}$ and $\mathcal{C}$ will be updated by gradient backpropped from $\mathcal{D}$, in other words, $\mathcal{S}$ and $\mathcal{C}$ act as auxiliary networks to make $\mathcal{D}$ a better discriminator in the discriminative iteration.
On the other hand, while in the generative iteration, $\mathcal{S}$ and $\mathcal{C}$ help to  make $\mathcal{G}$ a better generator to ``fool'' $\mathcal{D}$ by Eq. (\ref{eq-loss-G}).
From this point, both $\mathcal{S}$ and $\mathcal{C}$ of our approach are  adversarial for their function in the training process, together with $\mathcal{G}$ and $\mathcal{D}$.

However, if $\mathcal{D}$ was excluded, the rest three subnets will constitute  a conventional auto-encoder, for $\mathcal{G}$ to be the decoder while $\mathcal{S}$ together $\mathcal{C}$ to be the encoder's part. Thus make it posssible to add a pixel-wise reconstruction term to the loss function  in additional fooling the discriminator. 
We choose $L_1$ loss as an additional term to enhance the similarity meric, this is a highly popular choice used in many related auto-encoder GAN variants, such as $\alpha$-GAN, PPGN, and PPGN\cite{roscaVariationalApproachesAutoEncoding2017, zhuUnpairedImagetoImageTranslation2017, nguyenPlugPlayGenerative2016}.
Then the hybrid objective function for generative iteration can be  sumarized as:
\begin{equation}\label{eq-loss-hybrid}
    \begin{aligned}
        L(\mathcal{G}, \mathcal{D}^*) = &\mathbb{E}_{x'_{i, j}  \sim p_{\{\mathcal{G}(\bm{s_i}, \bm{c_j})\}}}[\log(1-\mathcal{D}(x'_{i, j}|\bm{s_i}^*, \bm{c_j}^*)) + \\
        & \lambda\|x'_{i, j} - x_{i, j}\|_1]
    \end{aligned}
\end{equation}
where $\lambda$ is a scale parameter that balances two terms in Eq. (\ref{eq-loss-hybrid}). We found the reconstruction term benificial to alleviate the mode-collapse degree of the adversarial model and accelerate the convergence of the min-max process in practice. 

\subsection{Optimization Algorithm}\label{sec-algorithm}
However, only with adversarial optimization on Eq. (\ref{eq-loss-D}) and Eq. (\ref{eq-loss-hybrid}) is not sufficient to achieve our task, it still won't ensure the features from $\mathcal{S}$ and $\mathcal{C}$ are reliable decoupled. This is because we haven't taken advantage of the feature co-sharing nature of glyph distribution yet, which is implied in Assumption \ref{asp-cs}.

As Eq. (\ref{eq-nets}) suggests, let $\{x_{i,j}\}$ and $\{x_{i, m}\}$ be two mini-batches paired in style, i.e., every sample pairs in these two mini-batches come from a same font set, 
we could also infer the corresponing style embeddings of the glyphs in mini-batch $\{x_{i,j}\}$ by feeding $\{x_{i, m}\}$ into $\mathcal{S}$. 
For the same reason, the content feature could also be infered with another mini-batch $\{x_{n, j}\}$, which paired with $\{x_{i,j}\}$ in character, by the deep network $\mathcal{C}$.

Therefore, in Eq. (\ref{eq-loss-D}), Eq. (\ref{eq-loss-G}) and Eq. (\ref{eq-loss-hybrid}), we use $\bm{s_i}$ and $\bm{c_j}$ extracted from $\{x_{i, m}\}$ and $\{x_{n, j}\}$, to get the point of equilibrium of the adversarially training.
Such a pairwise substitution extracting strategy prevents the well-trained $\mathcal{G}$ and $\mathcal{D}$ from relying on the low-level semantics of raw glyph images, thus forces the high level features to be decoded during training.
The overall flow of this substitution optimization is illustrated in Fig. \ref{fig-overall}, detailed algorithms are in the supplementary material.

During the optimization we keep two feature code stores of stlye and content subsects, i.e., $\mathbb{Z}_s$ and $\mathbb{Z}_c$, to feed the necessary known inputs as in Eq. (\ref{eq-netc}-\ref{eq-nets}) and Eq. (\ref{eq-loss-D}-\ref{eq-loss-G}).
As we want the content of $\mathbb{Z}_s$ and $\mathbb{Z}_c$ newest to the current model, 
after updating the parameters of  $\mathcal{S}$ and $\mathcal{C}$ in every iteration,
we will  randomly select two extra mini-batches to get new ranges of feature code.
To eliminate errors as much as possible, 
each feature embedding vector of the specific style or content will be averaged:
\begin{equation}
  \hat{\bm{s_i}} = \frac{1}{n_s}\sum_{k=1}^{n_s}{s_i^{(k)}}
  \label{eq-average-s}
\end{equation}
\begin{equation}
  \hat{\bm{c_j}} = \frac{1}{n_c}\sum_{l=1}^{n_c}{c_j^{(l)}}
  \label{eq-average-c}
\end{equation}
$n_s$, $n_c$ are the number of repeated glyph of font $i$ and character $j$ in two extra mini-batches, then $\mathbb{Z}_s$ and $\mathbb{Z}_c$ will be updated with these averaged codes.
But from the very begining, nothing is in $\mathbb{Z}_s$ and $\mathbb{Z}_c$, so we randomly inialize all known feature codes($\bm{s_n}^*$, $\bm{s_i}^*$, $\bm{c_m}^*$, $\bm{c_j}^*$ in Fig. \ref{fig-overall}) from $\mathcal{N}(0, \bm{I})$;
after the very first iteration, all the training mini-batch will be sampled in the range where $\mathbb{Z}_s$ and $\mathbb{Z}_c$ covers, so every necessary known inputs could be found in these two stores.

\section{Experiments}
\subsection{Data Preparation}
For this task, we build a glyph images ($128\times 128$ pixels) dataset from 60 regular Chinese fonts, 50 for traing while the rest for testing. 
Basically, these fonts are in the conventional Chinese font style categories known as  songti, heiti, kaiti and fangsong, but each varies in its weight, condensation, width, ornamentation.
Each font we selected includes at least the 6763 characters in GB-2312, which covers over 99\% of the characters of contemporary usage.
However, we believe that any glyph dataset that build on a sufficient number of Chinese text fonts covering thousands of characters would work in our proposed CocoAAN model, a small scale benchmark has validated this.

\subsection{Basic Model Setting}
All the four subnets are based on the general architecture of CNN models, details are sumarized in the supplementary material.
To stabilize the training of GAN model, we applied spectral normalization\cite{miyatoSpectralNormalizationGenerative2018} for the layers in discriminator, which will constrain the Lipschitz constant of the discriminator by restricting the spectral norm of each layer.
As the encoding networks play a vital role in discriminative iteration, a drastic escalation of parameters may cause the model unstable, we also used spectral normalization in $\mathcal{S}$ and $\mathcal{C}$.

In most other conditional 2D CNN model's implementation\cite{perarnauInvertibleConditionalGANs2016}, the 1D latent code is integrated into the input by expansion-concatenation operation, i.e., the latent is firstly expanded to the size of feature map, then concatenated into the input in channel. 
Such an incorporating method will bring a lot of invalid calculation in the subsequent convolution layer, since the newly concatenated feature maps were merely repetition of itself.

\begin{figure}[htbp]
  \centering
  \includegraphics[width=.8\linewidth]{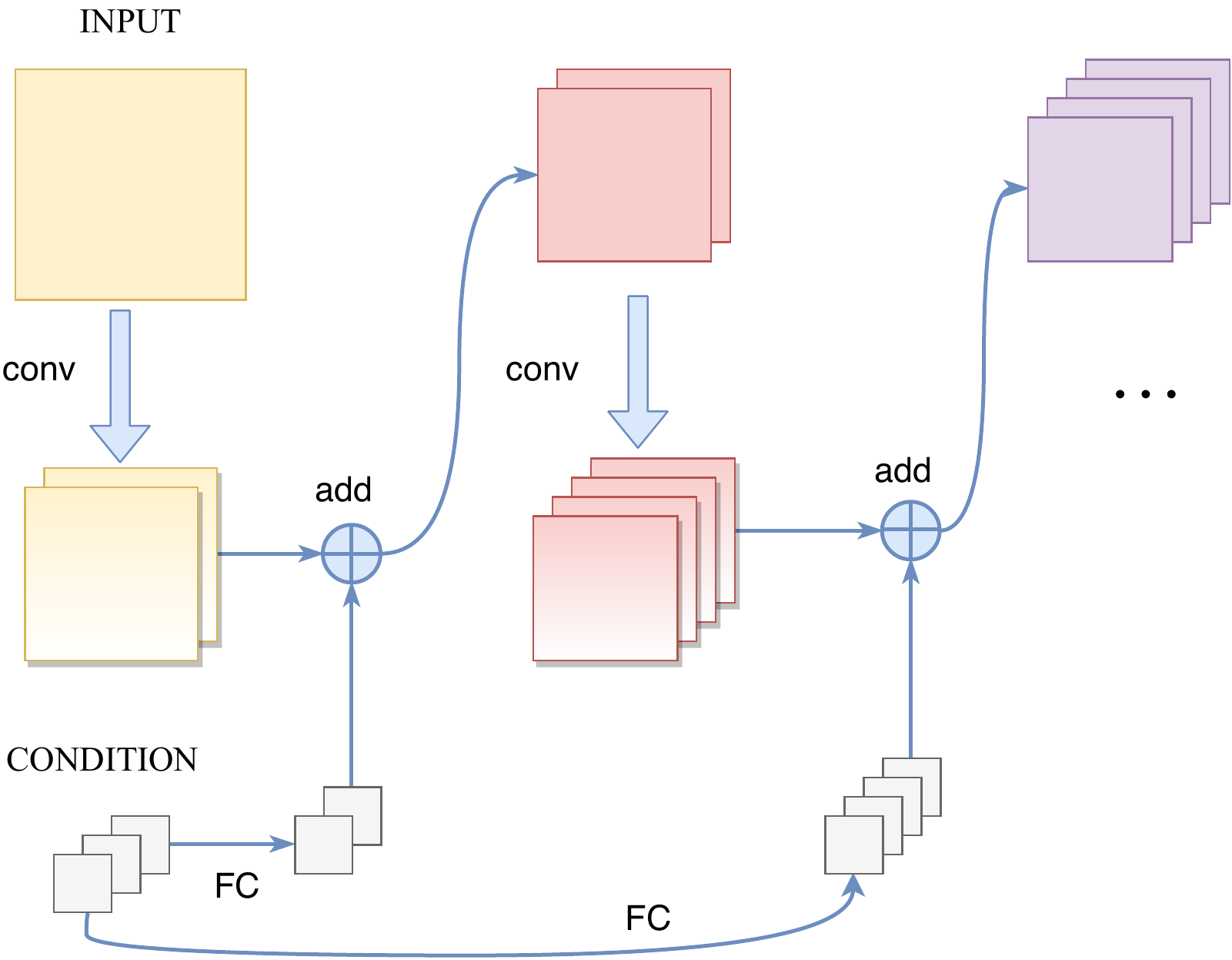}
  \caption{Our proposed Fully Connect-Add (FC-Add) operation.}
  \label{fig-condition-add}
\end{figure}

We introduce a new incorporating method for lattent code, the Fully Connect-Add (FC-Add) operation. 
The process is as follows (Fig. \ref{fig-condition-add}):
the conditional lattent code will firstly be resized to the the current input's channel shape by a fully connect layer, then added to the current input channelwisely.
For better conditional attribute learning, this FC-Add operation can be repeated several times at early stages  of downsampling networks.
As in our case, the very first three convolution layers of $\mathcal{S}$, $\mathcal{C}$, $\mathcal{D}$ are all followed with this modification.

No pre-processing was applied to training images besides scaling to the range of the tanh activation function $[-1, 1]$. All models were trained with Adam optimizer\cite{kingmaAdamMethodStochastic2014}  with $\beta_1=0$ and $\beta_2=0.9$.
By default, we use the the two-timescale update rule (TTUR) for the learning rate setting as it is proved to be effective in a regularized discriminator scenario\cite{heuselGANsTrainedTwo2017}.
Learning rate is set to 0.0001, 0.0001, 0.0002, 0.0004 for $\mathcal{S}$, $\mathcal{C}$, $\mathcal{G}$ and $\mathcal{D}$, respectively.
We set $\lambda = 10$ in Eq. (\ref{eq-loss-hybrid}) for all the experiments.

\subsection{Results}
\subsubsection{Reconstruction Results}
\begin{figure}[h!]
  \centering
  \subfigure[CocoAAN samples vs. ground gruth]{\label{fig-recon:a}
  \includegraphics[width=0.8\linewidth]{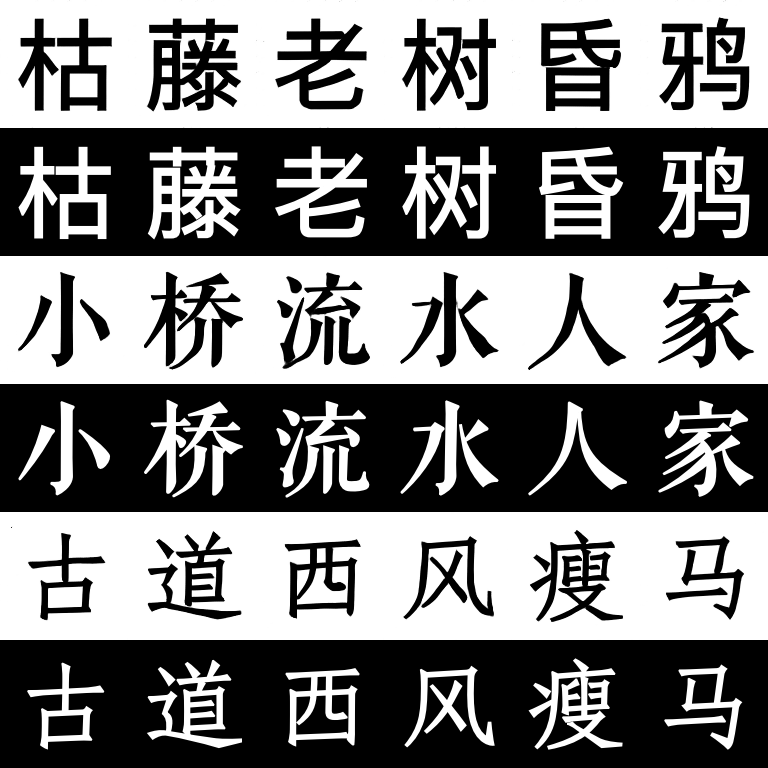}}
  \vfill  
  \subfigure[CocoAAN]{\label{fig-recon:b}
  \includegraphics[width=0.22\linewidth]{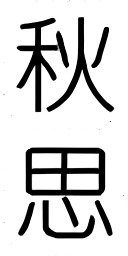}}
  \subfigure[VAE]{\label{fig-recon:c}
  \includegraphics[width=0.22\linewidth]{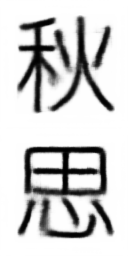}}
  \subfigure[GAN]{\label{fig-recon:d}
  \includegraphics[width=0.22\linewidth]{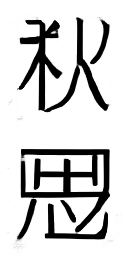}}
  \subfigure[Ground truth]{\label{fig-recon:e}
  \includegraphics[width=0.22\linewidth]{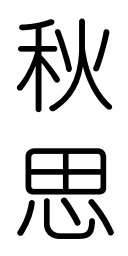}}  
  \caption{(a): Comparisons of reconstructed glyphs and the ground truth. Glyphs in odd lines are the reconstructed samples by our CocoAAN, and the adjacent background inversed lines are the ground truth. (b)-(e): Enlarged views of reconstructed glyphs of the same two Chinese characters from three models and the ground truth.}
  \label{fig-recon}
\end{figure}

Fig. \ref{fig-recon:a} shows parts of the reconstructed samples with the style and content features learned from the training dataset.
We can see results of CocoGAN are almost as real as the ground truth, the style and content details are both perfectly recovered except some little flaws. 
We also compare our results with those trained with VAE model and GAN model within the same iterations, while the settings and optimization  strategy keep the same. As shown in the rest parts of Fig. \ref{fig-recon}, VAE model produces blurry images, while the GAN (without L1 term as objective in generative iteration) method suffers from model collapse, our CocoAAN avoids these issues simultaneously.
Considered the experimental networks were not that deep, and the training images were merely in resolution ratio of $128 \times 128$, it demonstrates that our model has a capacity to precisely disentangle and restore domain-cross features of Chinese fonts.

\subsubsection{Style Learning}\label{sec-style-learn}
\begin{figure}[htbp]
  \centering
  \includegraphics[width=.8\linewidth]{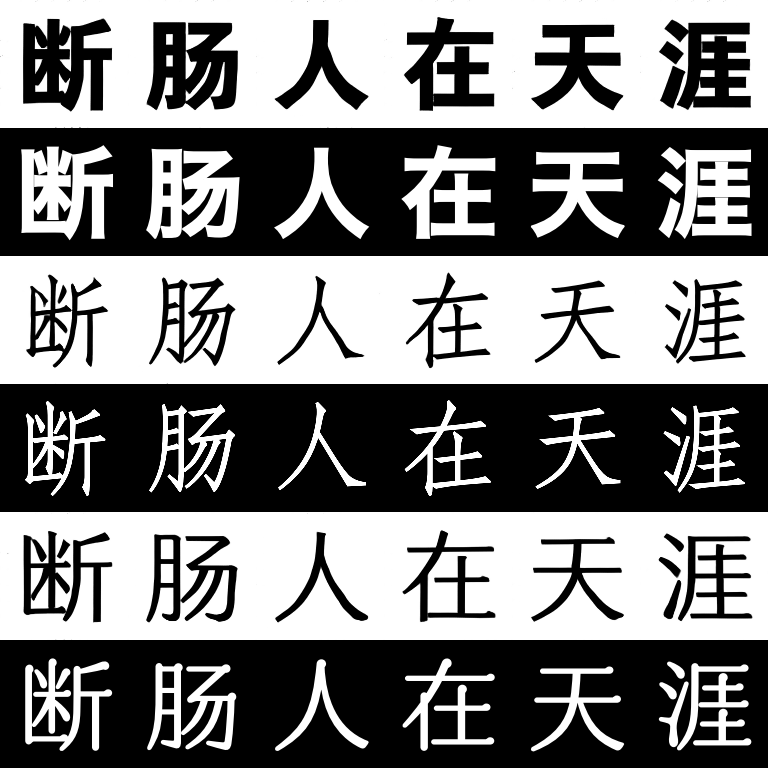}
  \caption{Comparisons of new stlye learning results and the ground truth. The black-background lines are the ground truth. The new style feature embeddings are infered from the test set.}
  \label{fig-stlye-learn}
\end{figure}
We test the capacity of learning new font style with the test dataset, based on the content feature embedding of the 6763 characters extracted from training dataset. Since each font possessed adequate glyphs, we averaged the newly learned style feature vector with a batchsize of 60 by Eq. (\ref{eq-average-s}).
Fig. \ref{fig-stlye-learn} lists samples drawn from the generator with the newly learned style features. 
Be critical, style details are not as fully acquired as the reconstruction results, e.g., glyphs in the 5th line of  Fig. \ref{fig-stlye-learn} totally missed the round serif in ground truth, which should be attributed to a lack of diversity of the training dataset.
But overall, our CocoAAN can generalize to fonts beyond the training dataset very well.

\begin{figure}[htbp]
  \centering
  \includegraphics[width=\linewidth]{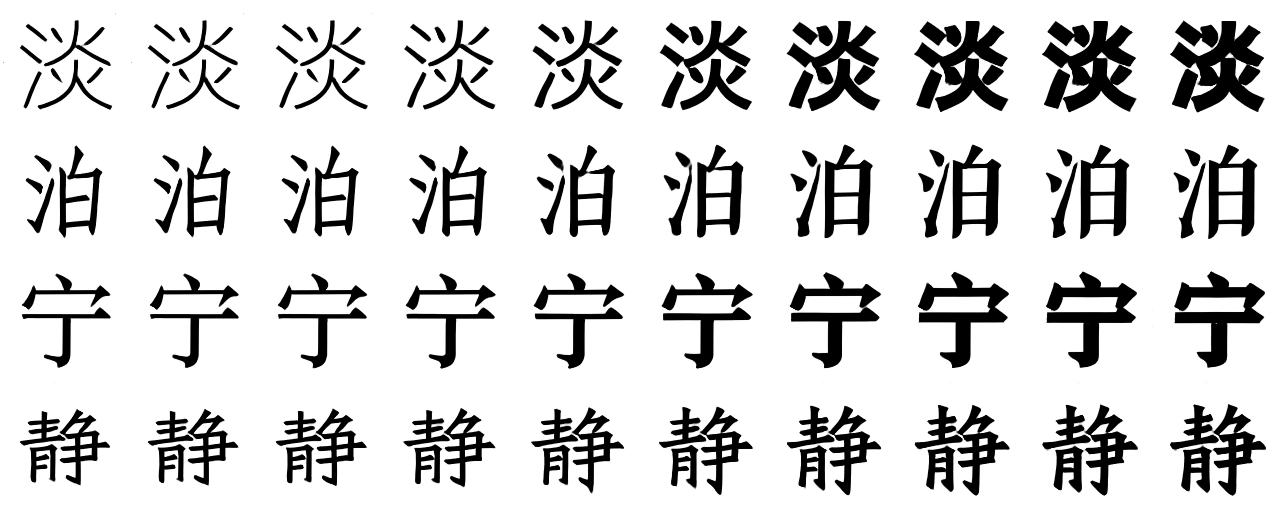}
  \caption{Linear interpolation results in style latent space. We intentionally choose interpolation between distinct styles, the left side tends to be thin while the right to be bold.}
  \label{fig-interp-s}
\end{figure}
Linear interpolation results in Fig. \ref{fig-interp-s} also show that the discinct styles can gradually transform along the style latent embeding, which proves that our encoding networks can learn a meaningful representation in the style latent space.

\subsubsection{Content Learning}
\begin{figure}[htbp]
  \centering
  \includegraphics[width=.8\linewidth]{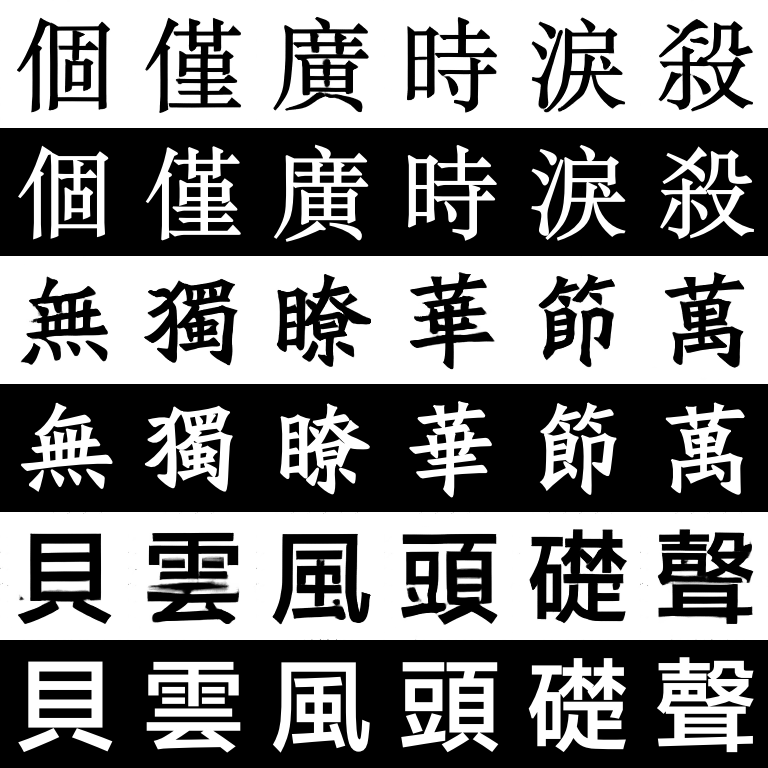}
  \caption{Comparisons of new content learning results and the ground truth. The black-background lines are the ground truth. The new content features are infered from glyphs of traditional Chinese characters, which are never appeared in the training set.}
  \label{fig-content-learn}
\end{figure}

Our CocoAAN also provides an automatic solution for encoding new character content features, instead of with manual encoding methods. 
In this test, we choose two conventional serif and sans-serif fonts as the input batch, which includes a range of GB-2312 unavailable characters, most of which are uncommon-used or more complicated traditional characters. As the manner in Sec. \ref{sec-style-learn}, each newly content embedding vector is averaged by Eq. (\ref{eq-average-c}). 
Fig. \ref{fig-content-learn} shows some result samples with the newly content feature embedding and other style feature in $\mathbb{Z}_s$.
Most of the characters can be well learned as a whole, 
but some results also showed an emergence of touched or overlapped strokes.

\begin{figure*}[htb]
  \centering
  \subfigure[]{\label{fig-tsne:a}
  \includegraphics[scale=1]{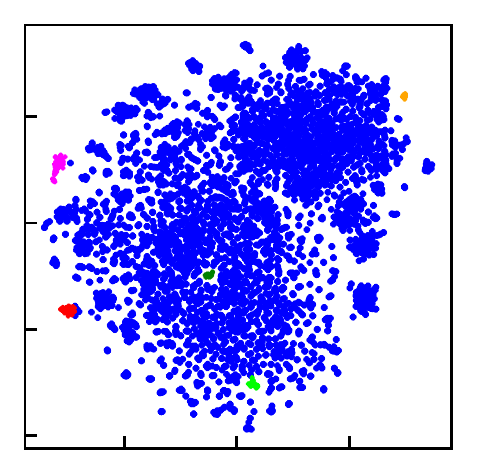}}
  \subfigure[]{\label{fig-tsne:b}
  \includegraphics[scale=1]{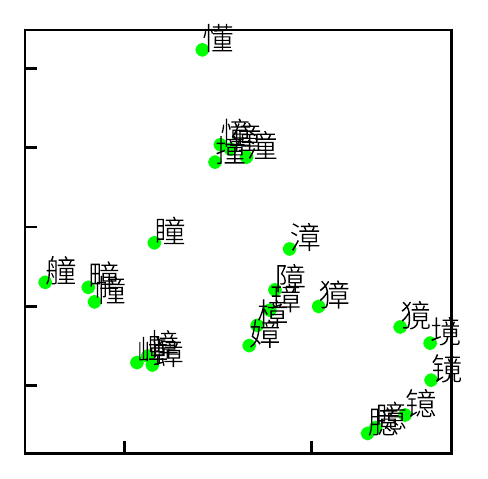}}
  \subfigure[]{\label{fig-tsne:c}
  \includegraphics[scale=1]{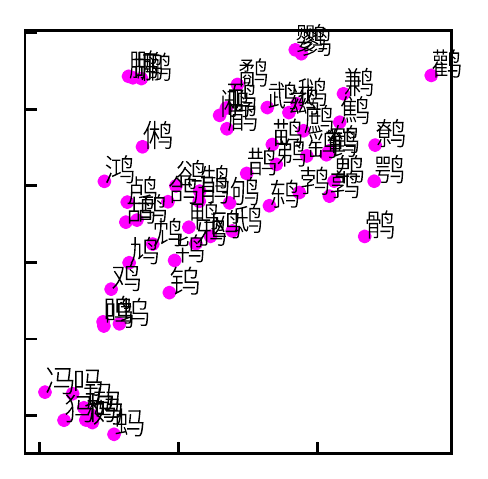}}
  \vfill
  \subfigure[]{\label{fig-tsne:d}
  \includegraphics[scale=1]{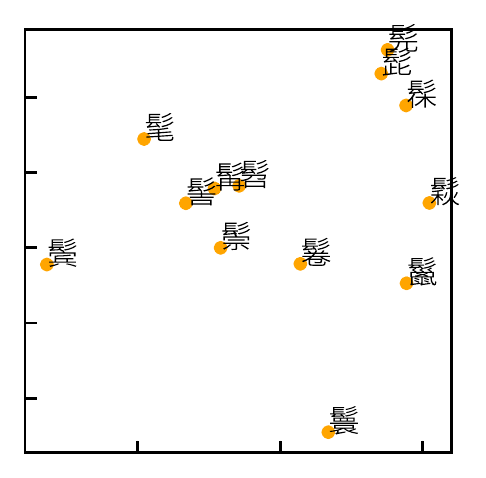}}
  \subfigure[]{\label{fig-tsne:e}
  \includegraphics[scale=1]{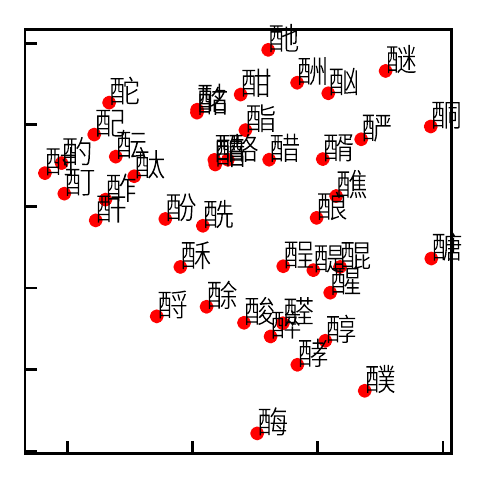}}
  \subfigure[]{\label{fig-tsne:f}
  \includegraphics[scale=1]{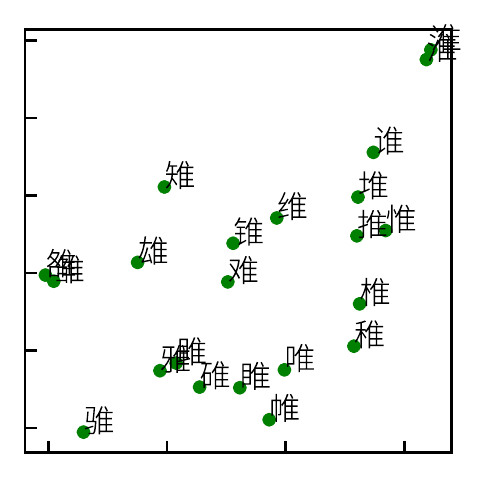}}
  \caption{(Best viewed in colors.) 2D Visualizations of the extracted hidden features by the content encoding network. (b)-(f) are  partial enlarged views of corresponding color regions in (a). Visualization is conducted with the t-SNE algorithm\cite{maatenVisualizingDataUsing2008}.}
  \label{fig-content-tsne}
\end{figure*}

We then visualize the content feature embedings of CocoAAN using the t-SNE method \cite{maatenVisualizingDataUsing2008} in Fig. \ref{fig-content-tsne}.
As shown in Fig. \ref{fig-content-tsne}(b)-(f), 
feature embeding of characters with the same or similar radicals are inclined to gather together. 
Without any hints of structure knowledge about the Chinese characters, our model learns to cluster the data in an unsupervised fashion automatically. Thus our framwork has successfully driven its content encoding part to capture reliable content representations in a meaningful way.

\section{Conclusions and future work}
In this paper, we propose a novel framework which jointly learns a generation network to achieve Chinese glyph synthesis and two encoding networks of the style and content related features
using an adversarial process. By the pairwise substitution optimization which based on the co-sharing nature of the style and content feature of glyphs, the model could get precise represntations from the style and content domains separately. This framwork shows an impressive ability to generalize to both other unseen fonts and  characters.

However, this framework focus more on printed Chinese fonts rather than the handwrittern ones, as the printed fonts are usually more rigidly designed and thus the style feature can be seen as a determined latent embedding which grounds the Assumption \ref{asp-cs}. For handwritten fonts more uncertainty is around, therefore regarding the style feature as a stochastic distribution may be a better choice. 
In this scenario, a varitional variant of the style inference network as VAE-GAN \cite{larsenAutoencodingPixelsUsing2015} may benefit to capture such an uncertain feature.

{\small
\bibliographystyle{ieee}
\bibliography{egbib}
}

\onecolumn
\begin{appendices}
\section{Training Algorithms}
  \begin{algorithm}
    \caption{Training algorithm of CocoAAN}
    \label{alg}
    \begin{algorithmic}
      \REQUIRE {Images of Chinese glyph sets: $X = \{x_{i, j}\}$}
      \ENSURE {The model parameters: $\theta_{\mathcal{C}}$, $\theta_{\mathcal{S}}$, $\theta_{\mathcal{D}}$, $\theta_{\mathcal{G}}$; Stores of feature of style and content subsects: $\mathbb{Z}_s$ and $\mathbb{Z}_c$}
      \STATE Randomly inialize $\theta_{\mathcal{C}}$, $\theta_{\mathcal{S}}$, $\theta_{\mathcal{D}}$, $\theta_{\mathcal{G}}$
  
      \FOR{number of training iterations}
      \STATE $\{x_{i, m}\}$, $\{x_{n, j}\}$, $\bm{s_i}^*$, $\bm{s_n}^*$, $\bm{c_j}^*$, $\bm{c_m}^*$ $\gets$ Preparation mini-batches and known inputs by algorithm \ref{alg-select-training}
      \STATE $\{x_{i, j}\} \gets X$, select 1 mini-batch sharing the same style with $\{x_{i, m}\}$ and same content with $\{x_{n, j}\}$
      \STATE $\bm{s_i}, \bm{c_j} \gets$ Get the feature code with $\mathcal{S}$ and $\mathcal{C}$
      \STATE $\theta_{\mathcal{S}}, \theta_{\mathcal{C}}, \theta_{\mathcal{D}} \gets$ Update parameters by ascending with $L(\mathcal{G}^*, \mathcal{D})$: 
      $$ \mathbb{E}_{x_{i, j} \sim p_{\mathrm{data}}}[\log \mathcal{D}(x_{i, j}|\bm{s_i}, \bm{c_j})] + 
      \mathbb{E}_{x'_{i, j}  \sim p_{\{G^*(\bm{s_i}^*, \bm{c_j}^*)\}}}[\log(1-\mathcal{D}(x'_{i, j}|\bm{s_i}, \bm{c_j}))] $$
      \STATE $\{x_{i, m}\}$, $\{x_{n, j}\}$, $\bm{s_n}^*$, $\bm{c_m}^*$ $\gets$ Preparation mini-batches and known inputs by algorithm \ref{alg-select-updating}
      \STATE Update $\mathbb{Z}_s$ by the output of $\mathcal{S}(x_{i, m}, \bm{c_m}^*)$
      \STATE Update $\mathbb{Z}_c$ by the output of $\mathcal{C}(x_{n, j}, \bm{s_n}^*)$
  
      \STATE $\{x_{i, m}\}$, $\{x_{n, j}\}$, $\bm{s_i}^*$, $\bm{s_n}^*$, $\bm{c_j}^*$, $\bm{c_m}^*$ $\gets$ Preparation mini-batches and known inputs by algorithm \ref{alg-select-training}
      \STATE $\bm{s_i}, \bm{c_j} \gets$ Get the feature code with $\mathcal{S}$ and $\mathcal{C}$
      \STATE $\theta_{\mathcal{S}}, \theta_{\mathcal{C}}, \theta_{\mathcal{D}} \gets$ Update parameters by descending with $L(\mathcal{G}, \mathcal{D}^*)$: 
      $$\mathbb{E}_{x'_{i, j}  \sim p_{\{\mathcal{G}(\bm{s_i}, \bm{c_j})\}}}[\log(1-\mathcal{D}(x'_{i, j}|\bm{s_i}^*, \bm{c_j}^*)) + \lambda\|x'_{i, j} - x_{i, j}\|_1]$$
      \STATE $\{x_{i, m}\}$, $\{x_{n, j}\}$, $\bm{s_n}^*$, $\bm{c_m}^*$ $\gets$ Preparation mini-batches and known inputs by algorithm \ref{alg-select-updating}
      \STATE Update $\mathbb{Z}_s$ by the output of $\mathcal{S}(x_{i, m}, \bm{c_m}^*)$
      \STATE Update $\mathbb{Z}_c$ by the output of $\mathcal{C}(x_{n, j}, \bm{s_n}^*)$
  
      \ENDFOR
    \end{algorithmic}
  \end{algorithm}

  \begin{algorithm}
    \caption{Selecting algorithm of training mini-batches}
    \label{alg-select-training}
    \begin{algorithmic}
      \REQUIRE {Images of Chinese glyph sets: $X = \{x_{i, j}\}$; Stores of feature of style and content subsects: $\mathbb{Z}_s$ and $\mathbb{Z}_c$; iteration number $i$}
      \ENSURE {Two mini-batches: $\{x_{i, m}\}$, $\{x_{n, j}\}$; The necessary knonwn inputs: $\bm{s_i}^*$, $\bm{s_n}^*$, $\bm{c_j}^*$, $\bm{c_m}^*$}
      \IF{first iteration}
      \STATE $\{x_{i, m}\}, \{x_{n, j}\} \gets X$, Randomly select 2 mini-batches from $X$
      \STATE $\bm{s_i}^*, \bm{s_n}^*, \bm{c_j}^*, \bm{c_m}^* \gets \mathcal{N}(0, \bm{I})$, Export necessary known inputs by random noise
      \ELSE
      \STATE $\{x_{i, m}\}, \{x_{n, j}\} \gets X$, Randomly select 2 mini-batches from $X$, in the range covered by both $\mathbb{Z}_s$ and $\mathbb{Z}_c$
      \STATE $\bm{s_i}^*, \bm{s_n}^* \gets \mathbb{Z}_s$, Export necessary known inputs according to corresponing style
      \STATE $\bm{c_j}^*, \bm{c_m}^* \gets \mathbb{Z}_c$, Export necessary known inputs according to corresponing content
      \ENDIF
    \end{algorithmic}
  \end{algorithm}
  
  \begin{algorithm}
    \caption{Selecting algorithm of updating mini-batches}
    \label{alg-select-updating}
    \begin{algorithmic}
      \REQUIRE {Images of Chinese glyph sets: $X = \{x_{i, j}\}$; Stores of feature of style and content subsects: $\mathbb{Z}_s$ and $\mathbb{Z}_c$; iteration number $i$}
      \ENSURE {Two mini-batches: $\{x_{i, m}\}$, $\{x_{n, j}\}$; The necessary knonwn inputs: $\bm{s_i}^*$, $\bm{s_n}^*$, $\bm{c_j}^*$, $\bm{c_m}^*$}
      \IF{first iteration}
      \STATE $\{x_{i, m}\}, \{x_{n, j}\} \gets X$, Randomly select 2 mini-batches from $X$
      \STATE $\bm{s_n}^*, \bm{c_m}^* \gets \mathcal{N}(0, \bm{I})$, Export necessary known inputs by random noise
      \ELSE
      \STATE $\{x_{i, m}\} \gets X$, Randomly select 1 mini-batch from $X$, in the range covered by $\mathbb{Z}_c$
      \STATE $\{x_{n, j}\} \gets X$, Randomly select 1 mini-batch from $X$, in the range covered by $\mathbb{Z}_s$
      \STATE $\bm{s_n}^* \gets \mathbb{Z}_s$, Export necessary known inputs according to corresponing style
      \STATE $\bm{c_m}^* \gets \mathbb{Z}_c$, Export necessary known inputs according to corresponing content
      \ENDIF
    \end{algorithmic}
  \end{algorithm}

\clearpage
\section{Network Architecture}\label{sec-net-details}
  \begin{table}[htbp]
    \centering
    \caption{model architecture of $\mathcal{G}$}
    \begin{tabular}{p{.4\linewidth}|p{.4\linewidth}}
      \toprule
      $\bm{s} \to 128\times 1\times 1$  & $\bm{c} \to 128\times 1 \times 1$ \\
      \hline
      $[4\times 4, 1]$, deconv $\to$ $512\times 4\times 4$, BN, ReLU &
      $[4\times 4, 1]$, deconv $\to$ $512\times 4\times 4$, BN, ReLU \\
      \hline
      \multicolumn{2}{c}{concatenate $\to 1024\times 4\times 4$} \\
      \hline
      \multicolumn{2}{c}{$[4\times 4, 2]$, deconv $\to 1024\times 8\times 8$, BN, ReLU} \\
      \multicolumn{2}{c}{$[4\times 4, 2]$, deconv $\to 512\times 16\times 16$, BN, ReLU} \\
      \multicolumn{2}{c}{$[4\times 4, 2]$, deconv $\to 256\times 32\times 32$, BN, ReLU} \\
      \multicolumn{2}{c}{$[4\times 4, 2]$, deconv $\to 128\times 64\times 64$, BN, ReLU} \\
      \multicolumn{2}{c}{$[4\times 4, 2]$, deconv $\to 1\times 128\times 128$, Tanh} \\
      \bottomrule
    \end{tabular}
  \end{table}

  \begin{table}[htbp]
    \centering
    \caption{model architecture of $\mathcal{S}$, $\mathcal{C}$}
    \begin{tabular}{p{.4\linewidth}|p{.4\linewidth}}
      \toprule
      input: $1\times 128\times 128$  & condition ($\bm{s}$ or $\bm{c}$): $\in \mathbb{R}^{128}$ \\
      \hline
      $[4\times 4, 2]$, SN conv $\to$ $64\times 64\times 64$ &
      FC condition $\to 64\times 1\times 1$ \\
      \hline
      \multicolumn{2}{c}{add $\to 64 \times 64 \times 64$, LeakyReLU} \\
      \hline
      $[4\times 4, 2]$, SN conv $\to$ $128\times 32\times 32$ &
      FC condition $\to 128\times 1\times 1$\\
      \hline
      \multicolumn{2}{c}{add $\to 128\times 32\times 32$} \\
      \hline
      $[4\times 4, 2]$, SN conv $\to$ $256\times 16\times 16$ &
      FC condition $\to 256\times 1\times 1$ \\
      \hline
      \multicolumn{2}{c}{add $\to 256\times 16\times 16$} \\
      \hline
      \multicolumn{2}{c}{$[4\times 4, 2]$, SN conv $\to 512\times 8\times 8$, LeakyReLU} \\
      \multicolumn{2}{c}{$[4\times 4, 2]$, SN conv $\to 1024\times 4\times 4$, LeakyReLU} \\
      \multicolumn{2}{c}{$[4\times 4, 2]$, SN conv $\to 128$ }\\
      \bottomrule
    \end{tabular}
  \end{table}

  \begin{table}[htbp]
    \centering
    \caption{model architecture of $\mathcal{D}$}
    \begin{tabular}{p{.4\linewidth}|p{.4\linewidth}}
      \toprule
      input: $1\times 128\times 128$  & condition ($\bm{s}$ and $\bm{c}$): $\in \mathbb{R}^{128}$ \\
      \hline
      $[4\times 4, 2]$, SN conv $\to$ $128\times 64\times 64$ &
      FC condition $\to 128\times 1\times 1$ \\
      \hline
      \multicolumn{2}{c}{add $\to 128 \times 64 \times 64$, LeakyReLU} \\
      \hline
      $[4\times 4, 2]$, SN conv $\to$ $256\times 32\times 32$ &
      FC condition $\to 256\times 1\times 1$\\
      \hline
      \multicolumn{2}{c}{add $\to 256\times 32\times 32$} \\
      \hline
      $[4\times 4, 2]$, SN conv $\to$ $512\times 16\times 16$ &
      FC condition $\to 512\times 1\times 1$ \\
      \hline
      \multicolumn{2}{c}{add $\to 512\times 16\times 16$} \\
      \hline
      \multicolumn{2}{c}{$[4\times 4, 2]$, SN conv $\to 512\times 8\times 8$, LeakyReLU} \\
      \multicolumn{2}{c}{$[4\times 4, 2]$, SN conv $\to 1024\times 4\times 4$, LeakyReLU} \\
      \hline
      \multicolumn{2}{c}{$[4\times 4, 2]$, SN conv $\to 1$} \\
      \bottomrule
    \end{tabular}
  \end{table}

\end{appendices}

\end{document}